\documentclass[10pt, a4paper]{article}
\usepackage{lrec2022} 
\usepackage{multibib}
\newcites{languageresource}{Language Resources}
\usepackage{graphicx}
\usepackage{tabularx}
\usepackage{soul}
\usepackage{titlesec}
\titleformat{\section}{\normalfont\large\bfseries\center}{\thesection.}{1em}{}
\titleformat{\subsection}{\normalfont\SmallTitleFont\bfseries\raggedright}{\thesubsection.}{1em}{}
\titleformat{\subsubsection}{\normalfont\normalsize\bfseries\raggedright}{\thesubsubsection.}{1em}{}
\renewcommand\thesection{\arabic{section}}
\renewcommand\thesubsection{\thesection.\arabic{subsection}}
\renewcommand\thesubsubsection{\thesubsection.\arabic{subsubsection}}

\usepackage{epstopdf}
\usepackage[utf8]{inputenc}

\usepackage{hyperref}
\hypersetup{
    colorlinks = true,
    allcolors = {blue}
}
\usepackage{xstring}

\usepackage{color}

\usepackage[acronym]{glossaries}

\makeglossaries

\newacronym{nlp}{NLP}{Natural Language Processing}
\newacronym{pie}{PIE}{Potential Idiomatic Expression}
\newacronym{mt}{MT}{Machine Translation}
\newacronym{wsd}{WSD}{word sense disambiguation}
\newacronym{mwe}{MWE}{Multi-Word Expression}
\newacronym{bnc}{BNC}{British National Corpus}
\newacronym{bert}{BERT}{Bidirectional Encoder Representations from Transformers}
\newacronym{pos}{PoS}{part of speech}
\newacronym{svm}{SVM}{Support Vector Machine}
\newacronym{mnb}{mNB}{multinomial Naive Bayes}
\newacronym{iaa}{IAA}{inter-annotator agreement}
\newacronym{sota}{SoTA}{state-of-the-art}
\newacronym{ukw}{UKWaC}{UK Web Pages}

\title{Potential Idiomatic Expression (PIE)-English: Corpus for Classes of Idioms}

\name{{\footnotesize Tosin Adewumi\thanks{Corresponding author- Tosin Adewumi, Machine Learning Group}, Roshanak Vadoodi*, Aparajita Tripathy, Konstantina Nikolaidou, Foteini Liwicki and Marcus Liwicki}}

\address{EISLAB, SRT, \\Exploration Geophysics, SBN*
\\
Luleå University of Technology, Sweden.
\\
firstname.lastname@ltu.se
}

\abstract{
We present a fairly large, \acrfull{pie} dataset for \acrfull{nlp} in English.
The challenges with \acrshort{nlp} systems with regards to tasks such as \acrfull{mt}, \acrfull{wsd} and information retrieval make it imperative to have a labelled idioms dataset with classes such as it is in this work.
To the best of the authors' knowledge, this is the first idioms corpus with classes of idioms beyond the literal and the general idioms classification.
In particular, the following classes are labelled in the dataset: metaphor, simile, euphemism, parallelism, personification, oxymoron, paradox, hyperbole, irony and literal.
We obtain an overall \acrfull{iaa} score, between two independent annotators, of 88.89\%.
Many past efforts have been limited in the corpus size and classes of samples but this dataset contains over 20,100 samples with almost 1,200 cases of idioms (with their meanings) from 10 classes (or senses).
The corpus may also be extended by researchers to meet specific needs.
The corpus has \acrfull{pos} tagging from the NLTK library.
Classification experiments performed on the corpus to obtain a baseline and comparison among three common models, including the \acrfull{sota} \acrshort{bert} model, give good results.
We also make publicly available the corpus and the relevant codes for working with it for \acrshort{nlp} tasks.
 \\ \newline 
}

\begin{document}

\maketitleabstract

\section{Introduction}
Idioms pose strong challenges to \acrshort{nlp} systems, whether with regards to tasks such as \acrshort{mt}, \acrshort{wsd}, information retrieval or metonymy resolution \cite{korkontzelos2013semeval}.
For example, in conversational systems, generating adequate responses depending on the idiom's class (for a user-input such as ``\textit{My wife kicked the bucket}") will benefit users of such systems.
This is because distinguishing the earlier example as an euphemism (a polite form of a hard expression), instead of just a general idiom, may elicit a sympathetic response from the conversational system, instead of a bland one.
More idiom examples (and their classes) in the dataset in this work are provided in section \ref{thecorpus}.
Also, classifying idioms into various classes has the potential benefit of automatic substitution of their literal meaning with \acrshort{mt}.

Idioms, which are part of figures of speech, are \acrfull{mwe} that have different meanings from the constituent meaning of the words \cite{quinn1993figures,drew1998figures}, though some draw a distinction between the two \cite{grant2004criteria}.
Not all \acrshort{mwe} are idioms.
An \acrshort{mwe} may be compositional, i.e. its meaning is predictable from the composite words \cite{diab2009verb}.
Research in this area is, therefore, important, especially since the use of idiomatic expressions is very common in spoken and written text \cite{lakoff2008metaphors,diab2009verb}.

Figures of speech are so diverse that a detailed evaluation is out of the scope of this work.
Indeed, figures of addition and subtraction create a complex but interesting collection \cite{quinn1993figures}.
Sometimes, idioms are not well-defined and classification of cases are not clear \cite{grant2004criteria,alm2003figures}.
Even single words can be expressed as metaphors \cite{lakoff2008metaphors,birke2006clustering}.
This fact makes distinguishing between figures of speech or idioms and literals quite a difficult challenge in some instances \cite{quinn1993figures}.
Previous work have focused on datasets without the actual classification of the senses of expressions beyond the literal and general idioms \cite{li2009classifier,cook2007pulling}.
Also, some of them have fewer than 10,000 samples \cite{sporleder2010idioms,li2009classifier,cook2007pulling}.
It is therefore imperative to have a fairly large dataset for neural networks training, given that more data increases the performance of neural network models. \cite{JMLR:v21:20-074,adewumi2019conversational,adewumi2020word2vec,adewumi2021sm}.

There are two usual approaches to idiom detection: type-based and tokens-in-context (or token-based) \cite{peng2015classifying,cook2007pulling,li2009classifier,sporleder2010idioms}.
The former attempts to distinguish if an expression can be used as an idiom while the latter relies on context for disambiguation between an idiom and its literal usage, as demonstrated in the SemEval semantic compositionality in context subtask \cite{korkontzelos2013semeval,sporleder2010idioms}.
This work focuses on the latter approach by presenting an annotated corpus.
The objectives, therefore, of this work are to create a high-quality corpus of potential idiomatic expressions in the English language and make it publicly available\footnote{github.com/tosingithub/idesk} for the \acrshort{nlp} research community.
This will contribute to advancing research in token-based idiom detection, which has enjoyed less attention in the past, compared to type-based.
Identification of fixed syntax (or static) idioms is much easier than those with inflections since exact phrasal match can be used.

The idioms corpus has almost 1,200 cases of idioms (with their meanings) (e.g. cold feet, kick the bucket, etc), 10 classes (or senses, including literal) and over 20,100 samples from, mainly, the \acrfull{bnc} with 96.9\% and about 3.1\% from \acrfull{ukw} \cite{ferraresi2008introducing}.
This is, possibly, the first idioms corpus with classes of idioms beyond the literal and general idioms classification.
The authors further carried out classification experiments on the corpus to obtain a baseline and comparison among three common models, including the \acrshort{bert} model.
The following sections include related work, methodology for creating the corpus, corpus details, experiments and the conclusion.

\section{Related Work}
There have been variations in the methods used in past efforts at creating idioms corpora.
Some corpora have less than 100 cases of idioms, less than 10,000 samples with few classes and without classification of the idioms \cite{sporleder2010idioms}.
Furthermore, labelled datasets for idioms in English are minimal.
Table 1 summarizes some of the related work, in comparison to the authors'.

The IDIX corpus, based on expressions from the \acrshort{bnc}, does not classify idioms, though annotation was more than the literal and non-literal alternatives \cite{sporleder2010idioms}.
They used Google search to ascertain how frequent each idiom is for the purpose of selection.
Their automatic extraction from the \acrshort{bnc} returned some erroneous results which were manually filtered out.
It contains 5,836 samples and 78 cases.
\newcite{li2009classifier} extracted 3,964 literal and non-literal expressions from the Gigaword corpus.
The expressions covered only 17 idiom cases.
Meanwhile, \newcite{cook2007pulling} selected 60 verb-noun construct (VNC) token expressions and extracted 100 sentences for each from the \acrshort{bnc}.
These were annotated using two native English speakers \cite{cook2007pulling}.

\newcite{diab2009verb} used \acrfull{svm} to perform binary classification into literal and idiomatic expressions on a subset of the VNC-Token.
The English SemEval-2013 dataset had over 4,350 samples \cite{korkontzelos2013semeval}.
The annotation did not include idiom classification but differentiated literal, figurative use or both, by using three crowd-workers per example.
It only contained idioms (from a manually-filtered list) that have their figurative and literal use, excluding those with only figurative use.

\newcite{saxena2020epie} introduced English Possible Idiomatic Expressions (EPIE) corpus, containing 25,206 samples of 717 idiom cases.
The dataset does not specify the number of literal samples and does not include idioms classification.
\newcite{haagsma2020magpie} generated potential idiomatic expressions in a recent work (MAGPIE) and annotated the dataset using only two main classes (idiomatic or literal), through crowdsourcing.
The idiomatic samples are 2.5 times more frequent than the literals, with 1,756 idiom cases and an average of 32 samples per case.
There are 126 cases with only one instance and 372 cases with less than 6 instances in the corpus, making it potentially difficult for neural networks to learn from the samples of such cases due to sample dearth.

Out of the two usual approaches to idiom detection (type-based and token-based) in the literature \cite{cook2007pulling,li2009classifier,sporleder2010idioms},
token-based detection is a more difficult task than semantic similarity of words and compositional phrases, as demonstrated by \newcite{korkontzelos2013semeval}, hence, detecting any of the multiple classes in an idioms dataset may be even more challenging.

There are various classes (or senses) of idioms, including metaphor, simile and paradox, among others \cite{alm2003figures}.
Tropes and Schemes, according to \newcite{alm2003figures}, are sub-categories of figures of speech.
Tropes have to do with variations in the use of lexemes and \acrshort{mwe}.
Schemes involve rhythmic repetitions of phoneme sequences, syntactic constructions, or words with similar senses.
A figure of speech becomes part of a language as an idiom when members of the community repeatedly use it.
The principles of idioms are similar across languages but actual examples are not comparable or identical across languages \cite{alm2003figures}.

\begin{table}[h]
\centering
\begin{tabular}{c|c|c|c}
\textbf{Dataset} &
\textbf{Cases} &
\textbf{Classes} & \textbf{Samples}
\\
\hline
PIE-English (ours) & 1,197 & 10 & 20,174
\\
\hline
IDIX & 78 & \footnotesize{NA*} & 5,836
\\
\hline
\textit{Li \& Sporleder} & 17 & 2 & 3,964
\\
\hline
MAGPIE & 1,756 & 2 & 56,192
\\
\hline
EPIE & 717 & \footnotesize{NA*} & 25,206
\\
\hline
\end{tabular}
\caption{\label{table:comp}Some datasets compared \footnotesize{(*NA: not available)}}
\end{table}

\section{Methodology}
We selected idioms from the dictionary by Easy Pace Learning\footnote{easypacelearning.com} in an alphabetical manner and samples were selected from the \acrshort{bnc} and \acrshort{ukw} based on the first to appear in both corpora.
Each sample contains 1 or 2 sentences, with the majority containing just 1.
The \acrshort{bnc} is a popular choice for text extraction for realistic samples across domains.
The \acrshort{bnc} is, however, relatively small, hence we relied also on the second corpus, \acrshort{ukw}, for further extraction when search results were less than the requirements (15 idiom samples and 21 for cases that have both idioms and literals).
Therefore, in each case, the number of samples were 22 for cases with literals and 16 for cases without literals (because of the included \acrshort{mwe}).
Six samples were decided to be the number of literal samples for each case that had both potential idiomatic expression and literal because the \acrshort{bnc} and \acrshort{ukw} sometimes had fewer or more literal samples, depending on the case.

Each of the 4 contributors (who are second/L2 English speakers) collected sample sentences of idioms and literals (where applicable) from the \acrshort{bnc}, based on identified idioms in the dictionary.
As a form of quality control, the entire corpus was reviewed by a near-native speaker.
This approach avoided common problems noticeable with crowd-sourcing methods, such as cheating the system or fatigue \cite{haagsma2020magpie}.
Although this approach is time-intensive, it also eliminates the problem noticeable with automatic extraction, such as duplicate sentences \cite{saxena2020epie} or false negatives/positives \cite{sporleder2010idioms}, for which manual effort may later be required.
This strategy gives high precision and recall to our total collection \cite{sporleder2010idioms}.

The contributors were given ample time for their task to mitigate against fatigue, which can be a common hindrance to quality in dataset creation.
We used the resources dedicated to the \acrshort{bnc} and other corpora\footnote{ phrasesinenglish.org/searchBNC.html \& \\ corpus.leeds.ac.uk/itweb/htdocs/Query.html} to extract the sentences.
The \acrshort{bnc} has 100M words while the \acrshort{ukw} has 2B words.
One of the benefits of these tools is the functionality for lemma-based search when searching for usage variants.
In a few cases, where less than 6 literal samples were available from both corpora, we used inflection to generate additional examples.
For example, \textit{"You need one to hold the ferret securely while the other ties the knot"} was inflected as \textit{"She needs one to hold the ferret securely while he ties the knot"}.
Two independent annotators were involved in this work.
Google search was used for cases in the dictionary that did not include classification and most of such came from The Free Dictionary\footnote{idioms.thefreedictionary.com}.

\section{The Corpus}
\label{thecorpus}
\begin{figure}[!htb]
   \begin{minipage}{.5\textwidth}
     \centering
     \includegraphics[width=1\linewidth]{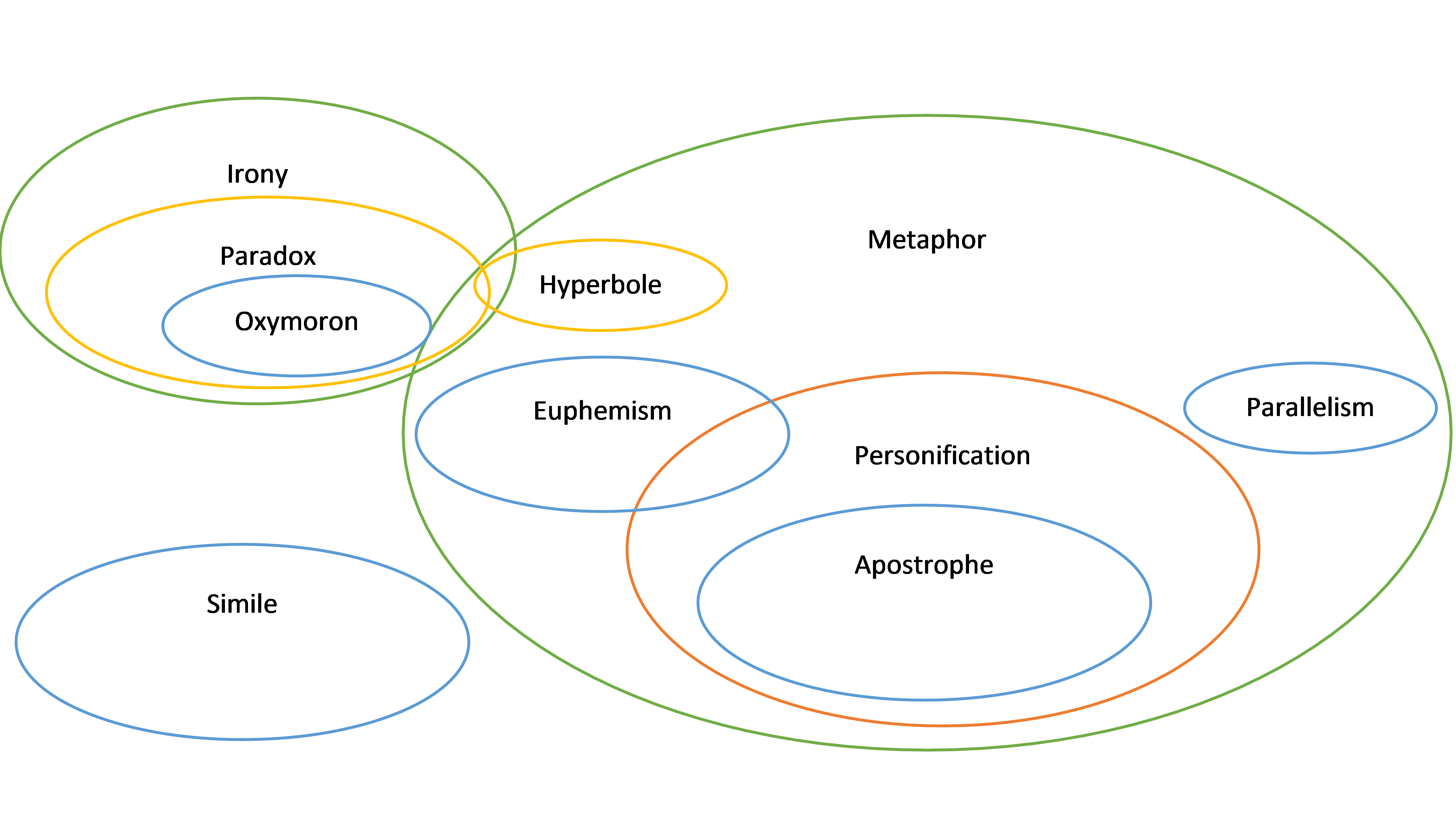}
     \caption{Classes of idioms \& their relationships}\label{Fig:idiomsset}
   \end{minipage}\hfill
\end{figure}

Idioms classification can sometimes overlap, as shown in Figure \ref{Fig:idiomsset}, and there is no general consensus on all the cases \cite{grant2004criteria,alm2003figures}.
Indeed, there have been different attempts at classifying idioms, including semantic, syntactic and functional classifications \cite{grant2004criteria,cowie1983oxford}.
It can be observed that a classification of a case or sample as personification also fulfills classification as metaphor, as it is also the case with euphemism.
Hence, the incident of two annotators with such different annotations does not imply they are wrong but that one is more specific.
Table \ref{table:meta} gives the distribution of the classes of samples.
The near-native speaker is responsible for annotation 1 in Table \ref{table:meta2}, based on their characteristics/guideline as discussed in this section, while the author of the dictionary is responsible for annotation 2.
A common approach for annotation is to have two or more annotators and determine their \acrshort{iaa} scores \cite{peng2015classifying}.
The overall \acrshort{iaa} score is 88.89\%.
Adjudication for the remaining 11.11\% cases for this dataset was to accept the classification guideline based on \newcite{alm2003figures}.
The \acrshort{iaa} score per class is the lower score between the two annotators, given in Table \ref{table:meta2}.

A metaphor uses a phenomenon or type of experience to outline something more general and abstract \cite{alm2003figures,lakoff2008metaphors}.
It describes something by comparing it with another dissimilar thing in an implicit manner.
This is unlike simile, which compares in an explicit manner.
Some other figures of speech sometimes overlap with metaphor and other idioms overlap with others.
Personification describes something not human as if it could feel, think or act in the same way humans could.
Examples of personification are metaphors also.
Hence, they form a subset (hyponym) of metaphors.
Apostrophe denotes direct, vocative addresses to entities that may not be factually present (and is a subset of personification) \cite{alm2003figures}.
Oxymoron is a contradictory combination of words or phrases.
They are meaningful in a paradoxical way and some examples can appear hyperbolic \cite{alm2003figures}.
Hyperbole is an exaggeration or overstatement.
This has the effect of startling or amusing the hearer. 
Figure \ref{Fig:idiomsset} is a diagram of the relationship among some classes of idioms, based on the authors' perception of the description by \newcite{alm2003figures}.

The idioms in the dataset are common in many English-speaking countries.
There is no restriction on the syntactic pattern of the idioms in the samples.
Our manual extraction approach from the base corpora increases the quality of the samples in the dataset, given that manual approaches appear to give more accurate results though demanding on time \cite{roh2019survey}.
Risks with data privacy are limited to what is provided in the base corpora (\acrshort{bnc} and \acrshort{ukw}).
Part of speech (\acrshort{pos}) tagging was performed using the natural language toolkit (NLTK) to process the original dataset \cite{loper2002nltk}.
Table \ref{table:columns} shows the columns in the corpus.
The corpus may also be extended by researchers to meet specific needs.
For example, by adding more samples for the cases from the \acrshort{bnc} or other reliable sources, adding more cases with their samples, or adding IOB tags for chunking, as another approach for training.

\begin{table}[h]
\centering
\begin{tabular}{c|c|c}
\textbf{Classes} & \textbf{\% of Samples} &
\textbf{Samples}
\\
\hline
Euphemism & 11.82 & 2,384
\\
\hline
Literal & 5.65 & 1,140
\\
\hline
Metaphor & 72.7 & 14,666
\\
\hline
Personification & 2.22 & 448
\\
\hline
Simile & 6.11 & 1,232
\\
\hline
Parallelism & 0.32 & 64
\\
\hline
Paradox & 0.56 & 112
\\
\hline
Hyperbole & 0.24 & 48
\\
\hline
Oxymoron & 0.24 & 48
\\
\hline
Irony & 0.16 & 32
\\
\hline
\textbf{Overall} & \textbf{100} &
\textbf{20,174}
\\
\hline
\end{tabular}
\caption{\label{table:meta}Distribution of samples of idioms/literals in the corpus}
\end{table}

\begin{table}[h]
\centering
\resizebox{\columnwidth}{!}{%
\begin{tabular}{c|c|c|c|c}
\textbf{Classes} &
\textbf{Annotation 1} &
\textbf{\%} &
\textbf{Annotation 2} &
\textbf{\%}
\\
\hline
Euphemism & 148 & 12.36 & 75 & 6.27
\\
\hline
Metaphor & 921 & 76.94 & 877 & 73.27
\\
\hline
Personification & 28 & 2.34 & 66 & 5.51
\\
\hline
Simile & 82 & 6.85 & 66 & 5.51
\\
\hline
Parallelism & 3 & 0.25 & 9 & 0.75
\\
\hline
Paradox & 6 & 0.5 & 19 & 1.59
\\
\hline
Hyperbole & 3 & 0.25 & 57 & 4.76
\\
\hline
Oxymoron & 4 & 0.33 & 9 & 0.75
\\
\hline
Irony & 2 & 0.17 & 19 & 1.59
\\
\hline
\textbf{Overall} & \textbf{1197} &
\textbf{100} &
\textbf{1197} & \textbf{100}
\\
\hline
\end{tabular}
}
\caption{\label{table:meta2}Annotation of classes of cases of idioms in the corpus}
\end{table}

\begin{table}[h]
\centering
\resizebox{\columnwidth}{!}{%
\begin{tabular}{c|c|c|c|c|c}
ID & Token &
PoS & class & meaning &
idiom+literal
\\
\end{tabular}
}
\caption{\label{table:columns}Columns in the corpus}
\end{table}

Examples of a sample per class in the corpus are given below. Each potential idiomatic expression in bracket represents a case.
\begin{enumerate}
    \item Metaphor (ring a bell): Those names ring a bell.
    \item Simile (as clear as a bell): it sounds as clear as a bell.
    \item Euphemism (go belly up): that Blogger could go belly up in the near future.
    \item Parallelism (day in, day out): that board was used day in day out.
    \item Personification (take time by the forelock): What I propose is to take time by the forelock.
    \item Oxymoron (a small fortune): a chest like this costs a small fortune if you can find one.
    \item Paradox (here today, gone tomorrow): he's a here today, gone tomorrow politician.
    \item Hyperbole (the back of beyond): Mhm. a voice came, from the back of beyond.
    \item Irony (pigs might fly): Pigs might fly, the paramedic muttered.
    \item Literal (ring a bell): They used to ring a bell up at the hotel.
\end{enumerate}

\subsection{Short data statement}
Data statements are important.
Failure to provide data statements could result in poor generalisability of results of trained models, harmful predictions, and failure of \acrshort{nlp} systems for certain groups \cite{bender2018data}.
It is beneficial to have a short version and a long, detailed version.
The long version of the \acrshort{pie}-English idioms corpus is provided in the appendix.

\begin{quote}
    \textbf{\textit{Short data statement for the \acrshort{pie}-English idioms corpus.}}\\
    This is the \acrfull{pie}-English idioms corpus for training and evaluating models in idiom identification.\\
    The licence for using this dataset comes under CC-BY 4.0.\\
    Total samples: 20,174\\
    There are 1,197 total cases of idioms and 10 classes.\\
    Total samples of euphemism (2,384), literal (1,140), metaphor (14,666), personification (448), simile (1,232), parallelism (64), paradox (112), hyperbole (48), oxymoron (48), and irony (32).
\end{quote}

\section{Experiments}
The data-split was done in a stratified way before being fed to the networks to address the class imbalance in the corpus.
This method ensures all classes are split in the same ratio among the training and dev (or validation) sets.
The split was 85:15 for the training and validation sets, respectively.
All experiments were performed on a shared cluster having Tesla V100 GPUs, though only one GPU was used in training the \acrshort{bert} model and the CPUs were used for the other classifiers.
Ubuntu 18 is the OS version of the cluster.

\subsection{Methodology}
The pre-processing involved lowering all cases and removing all html tags, if any, though none was found as the data was extracted manually and verified.
Furthermore, bad symbols and numbers were removed.
The training dataset is shuffled before training.
The following classifiers/models were experimented with to serve as some baseline and comparison: \acrfull{mnb} classifier, linear \acrshort{svm} and the \acrfull{bert} \cite{devlin2018bert}.
The authors used CountVectorizer as the matrix of token counts before transforming it into normalized TF-IDF representation and then feeding the \acrshort{mnb} and \acrshort{svm} classifiers.
\acrshort{bert}, however, uses WordPiece embeddings \cite{devlin2018bert}.
Batch size of 64 and total training epoch of 7 are used.
The \acrshort{svm} uses stochastic gradient descent (SGD) and hinge loss.
Its default regularization is l2.

\section{Results and Discussion}
Table \ref{table:resulttable} shows weighted average results obtained from the experiments, over three runs per model.
It will be observed that all three classifiers give results above what may be considered chance.
\acrshort{bert}, being a pre-trained, deep neural network model, performed best out of the three classifiers.
Table \ref{table:result2} shows that, despite the good results, the corpus can benefit from further improvement by addition of samples to the classes of idioms that have a low number.
This is because the classes with accuracy/F1 results close to zero are the ones with the least number of samples in the corpus.
Adding more samples to them should improve the results.
Regardless, there is strong performance in seven, out of the ten, classes in the corpus.

\begin{table}[h]
\centering
\begin{tabular}{c|c|c}
\textbf{Model} &
\textbf{Accuracy} &
\textbf{F1}
\\
\hline
mNB & 0.747 & 0.66
\\
\hline
SVM & 0.766 & 0.67
\\
\hline
BERT & \textbf{0.934} & \textbf{0.948}
\\
\hline
\end{tabular}
\caption{\label{table:resulttable}Weighted average results of classification of samples over all classes for the three models 
}
\end{table}

\begin{table}[h]
\centering
\begin{tabular}{c|c|c}
\textbf{Class} &
\textbf{Accuracy} &
\textbf{F1}
\\
\hline
Euphemism & 0.935 & 0.93
\\
\hline
Literal & 0.813 & 0.78
\\
\hline
Metaphor & 0.975 & 0.98
\\
\hline
Personification & 0.811 & 0.81
\\
\hline
Simile & 0.996 & 0.98
\\
\hline
Parallelism & 0.667 & 0.62
\\
\hline
Paradox & 0.725 & 0.82
\\
\hline
Hyperbole & 0.048 & 0.08
\\
\hline
Oxymoron & 0.095 & 0.15
\\
\hline
Irony & 0 & 0
\\
\hline
\end{tabular}
\caption{\label{table:result2}\acrshort{bert} average results over the classes of idioms}
\end{table}

\subsection{Error analysis}
\begin{figure}[h]
   \begin{minipage}{.5\textwidth}
     \centering
     \includegraphics[width=1\linewidth]{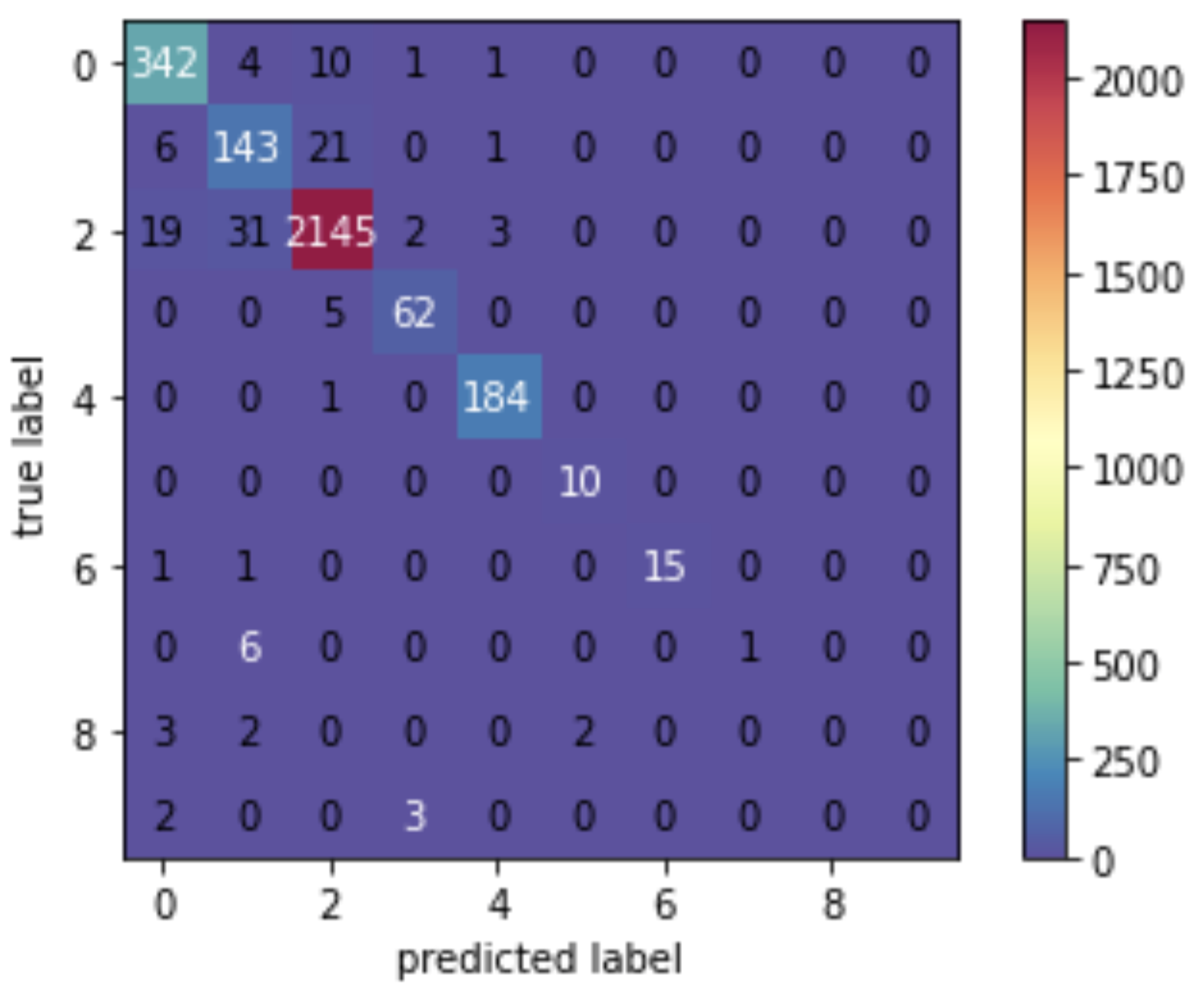}
     \caption{\acrshort{bert} Confusion Matrix \footnotesize{\\ 0=Euphemism, 1=Literal, 2=Metaphor, 3=Personification, 4=Simile, 5=Parallelism, 6=Paradox, 7=Hyperbole, 8=Oxymoron, 9=Irony.}}\label{Fig:conf_m}
   \end{minipage}\hfill
\end{figure}

Figure \ref{Fig:conf_m} presents error analysis through a confusion matrix, thereby providing more details about Table \ref{table:result2}.
It reveals the errors and successes made by the model.
We observe that most of the misclassification with metaphor are into literal, followed by euphemism.
Meanwhile, most of the misclassification with euphemism are into metaphor, possibly because metaphor is the largest class in the training set.

\section{Limitation}
A limitation of the \acrshort{pie}-English dataset, which seems inevitable, is the dominance of metaphors, since metaphors are the most common figures of speech \cite{bizzoni2017deep,grant2004criteria}.
Also, the corpus does not cover all possible idioms or figures of speech.

\section{Conclusion}
In this work, we address the challenge of non-availability of labelled idioms corpus with classes by creating one from the \acrshort{bnc} and the \acrshort{ukw} corpora.
It is possibly the first idioms corpus with classes of idioms beyond the literal and general idioms classification.
The dataset contains over 20,100 samples with almost 1,200 cases of idioms from 10 classes (or senses).
The dataset may also be extended to meet specific \acrshort{nlp} needs by researchers.
The authors performed classification on the corpus to obtain a baseline and comparison among three common models, including the \acrshort{bert} model \cite{devlin2018bert}.
Good results are obtained.
We also make publicly available the corpus and the relevant codes for working with it for \acrshort{nlp} tasks.

\section{Acknowledgements}
The authors wish to thank the anonymous reviewers for their valuable feedback.

%
%
%
%

\section{Bibliographical References}\label{reference}

\bibliographystyle{lrec2022-bib}
\bibliography{lrec2022-example}


\printglossary[type=\acronymtype]

\onecolumn

\newpage

\section*{Appendix}

\begin{table}[h]
\centering
\begin{tabular}{p{.17\textwidth}|p{.8\textwidth}}
\multicolumn{2}{l}{Data statement for the \acrshort{pie}-English idioms corpus for idiom identification.}
\\
 \hline
\textbf{} &
\textbf{Details}
\\
\hline
Curation rationale & Due to the unavailability of idioms dataset with more than the 2classes of literal \& general figurative speech classification, this dataset was created.
\\
\hline
Dataset language & English
\\
\hline
 & \textbf{Demographics of contributors}
 \\
\hline
No of contributors & 4
\\
\hline
Age & 42 | - | - | -
\\
\hline
Gender & Male | Female | Female | Female
\\
\hline
Language & L2 | L2 | L2 | L2
\\
\hline
 & \textbf{Demographics of annotators}
\\
\hline
No of annotators & 2
\\
\hline
  & \textbf{Annotator 1}
\\
\hline
Age & -
\\
\hline
Gender & Male
\\
\hline
Language & L2
\\
\hline
  & \textbf{Annotator 2}
\\
\hline
Age & -
\\
\hline
Gender & -
\\
\hline
Language & L2
\\
\hline
 & \textbf{Data characteristics}
\\
\hline
Total samples & 20,174
\\
\hline
Number of classes & 10
\\
\hline
Number of cases & 1,197 (e.g. \textit{“the nick of time"}, \textit{“a laugh a minute"})
\\
\hline
 & Total samples of euphemism (2,384), literal (1,140), metaphor (14,666), personification (448), simile (1,232), parallelism (64), paradox (112), hyperbole (48), oxymoron (48), and irony (32)
 \\
\hline
Base data & \acrshort{bnc} and \acrshort{ukw}.
\\
\hline
 & \textbf{Others}
\\
\hline
\acrshort{iaa} & 88.89\% (raw percentage)
\\
\hline
Licence & CC-BY 4.0.
\\
\hline
\end{tabular}
\caption{\label{appsvanalogy}}
\end{table}

\end{document}